%% file: neurips_2024.tex
\documentclass{article}

\PassOptionsToPackage{numbers, compress}{natbib}

\usepackage[preprint]{neurips_2024}

\usepackage[utf8]{inputenc} 
\usepackage[T1]{fontenc}    
\usepackage{hyperref}       
\usepackage{url}            
\usepackage{booktabs}       
\usepackage{amsfonts}       
\usepackage{nicefrac}       
\usepackage{microtype}      
\usepackage{xcolor}         
\usepackage{enumitem}
\usepackage{amsmath}
\usepackage{graphicx}
\usepackage{wrapfig}
\usepackage{amssymb}

\title{Physics3D: Learning Physical Properties of 3D Gaussians via Video Diffusion}

\author{
  Fangfu~Liu$^{1,}$\footnotemark[1]\quad Hanyang~Wang$^{1,}$\footnotemark[1]\quad Shunyu~Yao$^2$\quad Shengjun~Zhang$^1$\\ \textbf{Jie~Zhou}$^1$\quad \textbf{Yueqi~Duan}$^{1,}$\footnotemark[2]\\
  $^1$Tsinghua University \quad\quad $^2$Stanford University
}

\begin{document}
\newcommand{\eg}{\textit{e.g., }}
\newcommand{\ie}{\textit{i.e., }}

\maketitle

\renewcommand{\thefootnote}{\fnsymbol{footnote}}
\footnotetext[1]{Equal contribution. $\dagger$The corresponding author.}
\renewcommand{\thefootnote}{\arabic{footnote}}

\input{nips2024/0-abstract}
\input{nips2024/1-introduction}
\input{nips2024/2-related-work}
\input{nips2024/3-method}
\input{nips2024/4-experiments}
\input{nips2024/5-conclusion}


\medskip
{
\small
\bibliographystyle{plain}
\bibliography{reference}
}

\clearpage

\appendix
\input{nips2024/appendix}

\end{document}

%% file: nips2024/0-abstract.tex
\begin{abstract}
    In recent years, there has been rapid development in 3D generation models, opening up new possibilities for applications such as simulating the dynamic movements of 3D objects and customizing their behaviors. However, current 3D generative models tend to focus only on surface features such as color and shape, neglecting the inherent physical properties that govern the behavior of objects in the real world. To accurately simulate physics-aligned dynamics, it is essential to predict the physical properties of materials and incorporate them into the behavior prediction process. Nonetheless, predicting the diverse materials of real-world objects is still challenging due to the complex nature of their physical attributes. In this paper, we propose \textbf{Physics3D}, a novel method for learning various physical properties of 3D objects through a video diffusion model. Our approach involves designing a highly generalizable physical simulation system based on a viscoelastic material model, which enables us to simulate a wide range of materials with high-fidelity capabilities. Moreover, we distill the physical priors from a video diffusion model that contains more understanding of realistic object materials.
    Extensive experiments demonstrate the effectiveness of our method with both elastic and plastic materials. Physics3D shows great potential for bridging the gap between the physical world and virtual neural space, providing a better integration and application of realistic physical principles in virtual environments. Project page: \url{https://liuff19.github.io/Physics3D}.
\end{abstract}

%% file: nips2024/1-introduction.tex
\section{Introduction}\label{sec:intro}
In recent years, 3D computer vision has witnessed significant advancements, with researchers focusing on reconstructing or generating 3D assets~\cite{mildenhall2021nerf-eccv, kerbl20233d-gaussian, tang2023dreamgaussian, poole2022dreamfusion, hong2023lrm, li2024advances}, and even delving into the realm of 4D dynamics~\cite{ren2023dreamgaussian4d, alignling2023}. However, a common feature in these works is the emphasis on color space, which can be difficult in modeling realistic interactive dynamics, especially for applications in areas such as virtual/augmented reality and animation. Physics simulation is one of the most crucial methods to achieve a deeper understanding of the real world and enhance the effectiveness of interactive dynamics. Although conventional methods~\cite{stewart2000rigid, felippa2004introduction, kilian2005particle} describe behavior using continuous physical dynamic equations based on body-fixed mesh, they are usually difficult and time-consuming to generate complex 3D objects and suffer from highly nonlinear issues, large deformations or fracture-prone physical phenomena\cite{materialjiang2016}.

Powered by recent advances in implicit and explicit 3D representation techniques (\eg NeRF~\cite{mildenhall2021nerf-eccv} and 3D Gaussian Splatting~\cite{kerbl20233d-gaussian}), some researchers~\cite{physgaussianxie2023, li2022pac} have attempted to bridge the gap between rendering and simulation using the differentiable Material Point Method (MPM)~\cite{mlsmpmhu2018}, which enables efficient physical simulation driven by 3D particles (\ie 3D Gaussian kernels). PhysGaussian~\cite{physgaussianxie2023} extends the capabilities of 3D Gaussian kernels by incorporating physics-based attributes such as velocity, strain, elastic energy, and stress. This unified representation of material substance facilitates both simulation and rendering tasks. However, manual pre-design of physical parameters in PhysGaussian~\cite{physgaussianxie2023} remains a laborious and imprecise process. To avoid manually setting parameters, PhysDreamer~\cite{zhang2024physdreamer} leverages object dynamics learned from video generation models~\cite{svdblattmann2023, wang2023modelscope} to estimate a physical material parameter (\ie Young’s modulus). However, in practical applications, real-world objects often exhibit a complex composite nature, making it challenging for a simulation approach that relies solely on a single physical parameter to fully capture their dynamic behavior. This limits PhysDreamer to be primarily tailored for the simulation of hyper-elastic materials. Specifically, it encounters significant challenges when dealing with materials such as plastics, metals, and non-Newtonian fluids due to its heavy reliance on optimizing Young's modulus alone.
The inherent complexities in these materials surpass the capabilities of PhysDreamer, highlighting the need for a more comprehensive and robust approach that considers a broader range of physical properties for accurate and effective simulation.

In this paper, we propose \textbf{Physics3D}, a generalizable physical simulation system to learn various physical properties of 3D objects. Given a 3D Gaussian representation, we first expand the dimension of the physical parameters to capture both elasticity and viscosity. Then we design a viscoelastic Material Point Method (MPM) to simulate 3D dynamics. Through the simulation process, we decompose the deformation gradient into two separate components and calculate them independently to contribute to the overall force. Finally, leveraging the capabilities of the differentiable MPM, we optimize the physical parameters via the Score Distillation Sampling (SDS)~\cite{poole2022dreamfusion} strategy to distill physical priors from the video diffusion model. Iterating the MPM process and SDS optimization, Physics3D achieves high-fidelity and realistic performance in a wide range of materials. Extensive experiments demonstrate the efficacy and superiority of our proposed Physics3D over existing methods. In summary, our key contributions are as follows.
\begin{itemize}[leftmargin=*]

\item We propose a novel generalizable physical simulation system called Physics3D, which is capable of learning diverse material 3D dynamics in a physically plausible manner.

\item We introduce a viscoelastic Material Point Method to simulate both the viscosity and elasticity of 3D dynamics, enabling support of various material simulations. 

\item We design a learnable internal filling strategy to optimize part of 3D Gaussians and utilize the Score Distillation Sampling to optimize physical parameters from the video diffusion model. 

\item Experiments show Physics3D is effective in creating high-fidelity and realistic 3D dynamics, ready for various interactions across users and objects in the future.

\end{itemize}

\begin{figure*}[!t]
    \centering
    \includegraphics[width=1\linewidth]{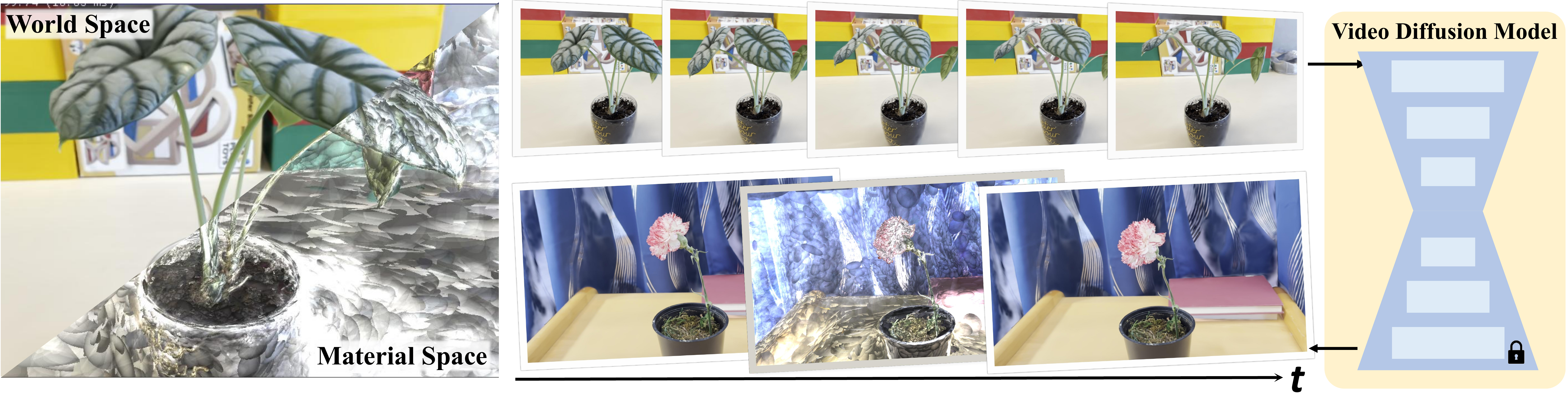}
    \vspace{-0.6cm}
    \caption{Physics3D is a unified simulation-rendering pipeline based on 3D Gaussians, which learn physics dynamics from video diffusion model.}
    \label{fig:gaussian_results}
    \vspace{-0.3cm}
\end{figure*}

%% file: nips2024/2-related-work.tex
\section{Related Work}

\textbf{Dynamic 3D representations.} Rapid advancements in static 3D representations have sparked interest in incorporating temporal dynamics into the 3D modeling of dynamic objects and scenes. Various explicit or hybrid representation techniques have demonstrated impressive outcomes, including planar decomposition for 4D space-time grids~\cite{cao2023hexplane, shao2023tensor4d, fridovich2023k}, the utilization of NeRF representation~\cite{li2022pac, pumarola2021d, gao2021dynamic}, and alternative structural approaches~\cite{turki2023suds, abou2024particlenerf, fang2022fast}. Recently, 3D Gaussian Splatting~\cite{kerbl20233d-gaussian} has revolutionized the representation via its outstanding rendering efficiency and high-quality results. Efforts have been made to extend static 3D Gaussians into dynamic versions, yielding promising results. Dynamic 3D Gaussians~\cite{luiten2023dynamic} refine per-frame Gaussian Splatting through dynamic regularizations and shared properties such as size, color, and opacity. Similarly, the concept of 4D Gaussian Splatting~\cite{wu20234d, yang2023deformable} employs a deformation network to anticipate time-dependent positional, scaling, and rotational deformations. In addition, DreamGaussian4D~\cite{ren2023dreamgaussian4d} learns motion from image-conditioned generated videos~\cite{blattmann2023stable-video-diffusion}, enabling more controllable and diverse 3D motion representations. 

\textbf{Viscoelastic materials.} In the realm of computer graphics and animation, there has been significant interest in accurately simulating the behavior of nonrigid objects and their interactions with physical environments. Conventional elastic models~\cite{terzopoulos1987elastically, zong2023neural} predicated on Hooke's law~\cite{rychlewski1984hooke-law} are the cornerstone for simulating the deformation of objects. These models are effective in representing materials that exhibit perfectly elastic behavior, returning to their original shape after the applied force is removed. However, real-world materials often exhibit more complex behaviors that cannot be captured by simple elastic models. The introduction of viscoelastic materials in computer graphics~\cite{terzopoulos1988modeling} has expanded the range of simulated material behaviors. Viscoelastic materials combine the characteristics of both viscous fluids and elastic solids, leading to time-dependent deformations under constant stress, a phenomenon known as creep~\cite{christensen2003theory}. Compared with elastic models, viscoelastic models offer a more versatile framework for animating nonrigid objects. They can simulate the slow restoration of a material's shape after the cessation of force, as well as the permanent deformation that occurs due to prolonged stress. 

\textbf{Video generation models.} With the emergence of models like Sora~\cite{sora2024}, the field of video generation~\cite{phenakivillegas2022, nuwawu2022, lumiere2024, alignyourlatents2023} has drawn significant attention. These powerful video models~\cite{videopoetkondratyuk2023, makeavideo2022, imagenvideo2022, cogvideohong2022} are typically trained on extensive datasets of high-quality video content. Sora~\cite{sora2024} is capable of producing minute-long videos with realistic motions and consistent viewpoints. Furthermore, some large-scale video models, such as Sora~\cite{sora2024}, can even support physically plausible effects. Inspired by these capabilities, we aim to distill the physical principles observed in videos and apply them to our static 3D objects, thereby achieving more realistic and physically accurate results. In our framework, we choose Stable Video Diffusion~\cite{svdblattmann2023} to optimize our physical properties.

%% file: nips2024/3-method.tex
\section{Problem Formulation}
Given a static representation through 3D Gaussians, our goal is to estimate the physical attributes of each Gaussian particle and generate physics-plausible motions by organizing the interaction of force and velocity among these particles. These physical properties include mass ($m$), Young’s modulus ($E$), and Poisson’s ratio ($\nu$). Young's modulus and Poisson's ratio controls dynamics of elastic objects. For example, with a fix amount of external force applied, system will higher Young's modulus will have smaller deformation.

However, only modeling the property of elastic objects is inadequate for recovering diverse physics with heterogeneous materials in real-world applications, which significantly limits the recent work like PhysDreamer~\cite{zhang2024physdreamer}. For example, they usually suffer from complex mixed materials, especially in scenarios of rapid deformation where viscosity emerges as a significant factor in dynamics. Therefore, our key insight is to build a more comprehensive physics model that includes additional parameters, notably viscosity, to enrich the descriptive capacity for real-world objects especially in inelastic scenarios. With a closer look at the viscosity coefficient $\nu_v$ and $\nu_d$, it serves as a viscous damper to exert resistance against rapid deformation, thus enhancing the model’s fidelity in capturing viscoelasticity.

To model viscoelastic stresses with physical fidelity, we explore continuum mechanics, where Lamé constants (also referred to as Lamé coefficients or parameters), denoted by $\lambda$ and $\mu$, emerge as pertinent material-related quantities within the strain-stress relationship, while viscosity coefficient $\nu_v$ and $\nu_d$ governs the viscous dynamics. Conventionally, $\lambda$ and $\mu$ are respectively denoted as Lamé's first and second parameters. Their significance varies in different contexts. For example, in fluid dynamics, $\mu$ is related to the dynamic viscosity of fluids while in elastic environments, Lamé parameters $\lambda$ and $\mu$ intertwine with Young’s modulus ($E$) and Poisson’s ratio ($\nu$) via a specific relationship. On the other hand, viscosity coefficient correct the elastic strain-stress relation by dynamically controlling the relation to account for viscosity.
Consequently, our framework pivots towards the estimation of the viscosity coefficient ($\nu_v$ and $\nu_d$) and the two Lamé parameters, $\lambda$, and $\mu$. As for other physical attributes, we align with the conventional methods~\cite{zhang2024physdreamer, physgaussianxie2023}, where the particle mass ($m_p$) is pre-calculated as the product of a constant density ($\rho$) and the particle volume ($V_p$), and Poisson’s ratio is constant across the object.

\begin{figure*}[!t]
    \centering
    \includegraphics[width=1\linewidth]{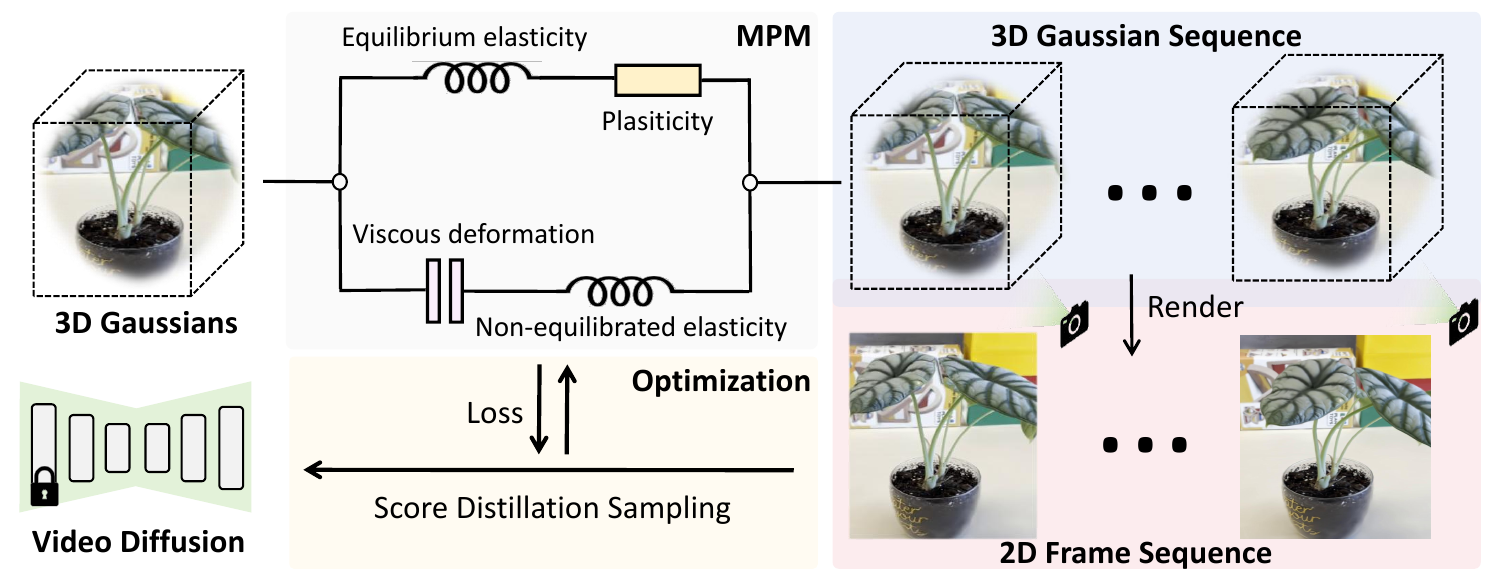}
    \caption{\textbf{Pipeline of Physics3D}. Given an object represented as 3D Gaussians, We first simulate it using the Material Point Method (MPM). The simulation comprises two distinct components: an elastoplastic part and a viscoelastic part. These components operate independently to calculate the individual stresses within the object and are combined to determine the overall stress. After the simulation, a series of Gaussians with varying orientations are generated, reflecting the dynamic evolution of the scene. Then, we render these Gaussians from a fixed viewpoint to produce a sequence of video frames. Finally, we utilize a pretrained video diffusion model with Score Distillation Sampling (SDS) strategy to iteratively optimize physical parameters.}
    \label{fig:pipeline}
\end{figure*}

\section{Method}\label{sec:method}
In this section, we introduce our method, \textit{i.e.}, Physics3D, for learning the dynamics of multi-material 3D systems with physical alignment. Our goal is to estimate the various physical properties of 3D objects. Building upon this, we first review the theory of three foundational techniques (referenced in Sec.~\ref{subsec:preliminary}) that form the backbone of our algorithm. Then we introduce our physical property estimation framework (Sec.~\ref{subsec:physcial_properties}) and describe our particle-based simulation process. Finally, we present the optimization process (Sec.~\ref{subsec:optimization}) to further improve the accuracy and efficiency. An overview of our framework is depicted in Figure~\ref{fig:pipeline}.

\subsection{Preliminary}\label{subsec:preliminary}

\begin{wrapfigure}{r}{0.35\linewidth}
    \vspace{-4.5em}
    \includegraphics[width=1\linewidth]{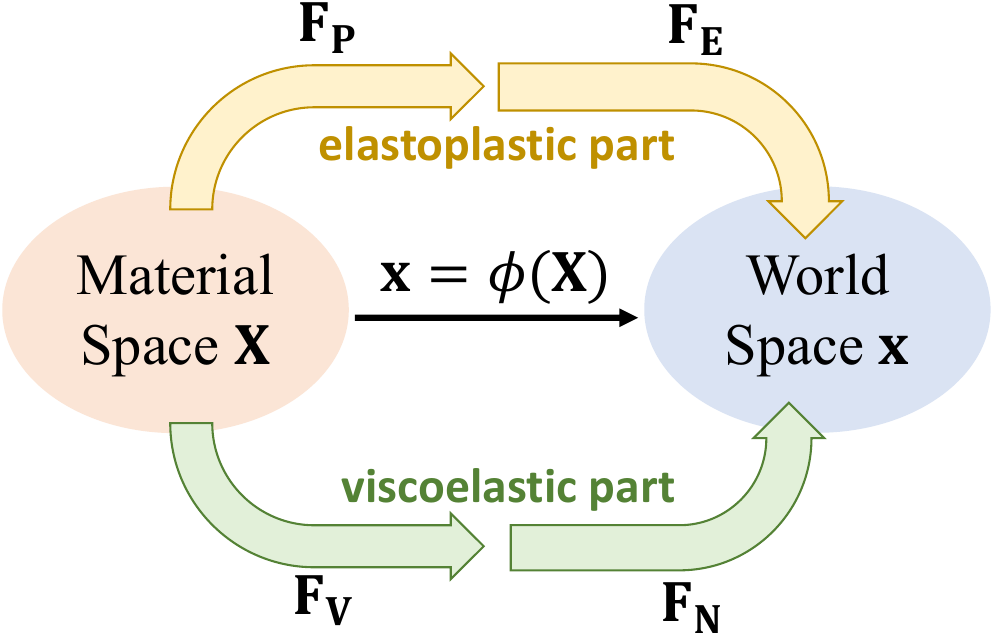}
    \vspace{-2em}
    \caption{Elastoplastic and viscoelastic decomposition.}
    \label{fig:decomposition}
    \vspace{-1em}
\end{wrapfigure}

\textbf{Continuum Mechanics} describes motions by a deformation map $\mathbf{x}=\phi(\mathbf{X},t)$ from the material space $\Omega^0$ (with coordinate $\mathbf{X}$ ) to the world space $\Omega^n$ (with coordinate $\mathbf{x}$). The deformation gradient $\mathbf{F} = \frac{\partial \phi}{\partial \mathbf{X}} (\mathbf{X}, t)$ measures local rotation and strain~\cite{bonet1997nonlinear}. We consider viscoelastic materials where we have two components~\cite{govindjee1997presentation}, the elastoplastic component $\mathbf{F_E} \mathbf{F_P}$ and the viscoelastic component $\mathbf{F_N} \mathbf{F_V}$. They are in parallel combination shown in Figure~\ref{fig:decomposition}, and formulated as: 
\begin{equation}
\label{eq:decomposition}
    \mathbf{F} = \mathbf{F_E} \mathbf{F_P} = \mathbf{F_N} \mathbf{F_V}.
\end{equation}
In this work, we take $\mathbf{F_P}=\mathbf{I}$ for convenience. Intuitively, we model our material with two different compounds in parallel connection as they deform in the same way thus having the same total strain. However, only the elastic components $\mathbf{F_E}$ and $\mathbf{F_N}$ contributes to the internal stress $\boldsymbol{\sigma_E}$ and $\boldsymbol{\sigma_N}$. 

We can now evolve the system with dynamical equations. Denoting the velocity field with $\mathbf{v}(\mathbf{x}, t)$ and density field with $\rho(\mathbf{x}, t)$, the conservation of momentum and conservation of mass~\cite{germain1998functional} is given by:
\begin{equation}\label{EOM}
    \rho \frac{D \mathbf{v} }{D t} = \nabla \cdot \boldsymbol{\sigma} + \mathbf{f},  \quad
    \frac{D\rho}{Dt} + \rho \nabla \cdot \mathbf{v} = 0,
\end{equation}
where $\mathbf{f}$ denotes an external force, $\boldsymbol{\sigma}=\boldsymbol{\sigma_E}+\boldsymbol{\sigma_N}$ is the total internal stress. We will also need to update the strain tensor after updating the material point, and we refer to the Appendix for more details.

\textbf{Material Point Method
(MPM)}~\cite{sulsky1995application, materialjiang2016, mlsmpmhu2018} discretizes the material into deformable particles and employs particles to monitor the complete history of strain and stress states while relying on a background grid for precisely evaluating derivatives during force computations. This methodology has demonstrated its efficacy in simulating diverse materials~\cite{apicjiang2015, druckerklar2016, materialstomakhin2013} and is proved to be capable of simulating some viscoelastic and viscoplastic materials~\cite{ram2015material, yue2015continuum}. MPM operates in a particle-to-grid (P2G) and grid-to-particle (G2P) transfer loop. In P2G process, MPM transfers mass and momentum from particles to grids:
\begin{equation}
    m^n_i= \sum_{p} w^n_{ip}m_p,  \quad
    m^n_i \mathbf{v^n_i}= \sum_{p} w^n_{ip}m_p (\mathbf{v^n_p} + C^n_p(\mathbf{x_i} - \mathbf{x^n_p})),
\end{equation}
Here $i$ and $p$ represent the fields on the Eulerian grid and the Lagrangian particles respectively. Each particle $p$ carries a set of properties including volume $V_p$, mass $m_p$, position $\mathbf{x_p^n}$, velocity $\mathbf{v_p^n}$, deformation gradient $\mathbf{F_p^n}$ and affine momentum $C_p^n$ at time $t_n$. $w^n_{ip}$ is the B-spline kernel defined on $i$-th grid evaluated at $\mathbf{x_p^n}$. After P2G, the transferred grids can be updated as:
\begin{equation}
    \mathbf{v_i^{n+1}} = \mathbf{v^n_i} - \frac{\Delta t}{m_i} \sum_{p} \tau^n_p \nabla w^n_{ip}V_p^0 + \Delta t \mathbf{g},
\end{equation}

here $\mathbf{g}$ represents the acceleration due to gravity. Then G2P transfers velocities back to particles and updates particle states $\tau$ (\ie Kirchhoff tensor).
\begin{equation}\label{eq:tensor}
    \tau_p^{n+1} = \boldsymbol{\tau}(\mathbf{F_E^{n+1}}, \mathbf{F_N^{n+1}}),
\end{equation}

where $\mathbf{F_E^{n+1}}$ and $\mathbf{F_E^{n+1}}$ are two parts of strain tensor. We will provide a detailed introduction in Sec.~\ref{subsec:physcial_properties}. In this work, we integrate the physical properties of viscoelastic materials into the Material Point Method, thereby enhancing the generalization capabilities of MPM. This implementation enables the simulation for a wide range of materials commonly found in the real world, including various inelastic materials. 

\textbf{Score Distillation Sampling (SDS)} is first introduced by DreamFusion~\cite{poole2022dreamfusion}, which distills the 3D knowledge from large 2D pretrain model~\cite{saharia2022photorealistic-imagen}. For a set of particles parameterized by physical properties ${\theta}$, its rendering $\mathbf{x}$ can be obtained by $\mathbf{x} = g({\theta})$ where $g$ is a differentiable renderer. SDS calculates the gradients of physical parameters ${\theta}$ by,
\begin{equation}
\label{eq:sds}
    \quad\nabla_{{\theta}}\mathcal{L}_{\text{SDS}}(\phi,\mathbf{x=g({\theta)}})=\mathbb{E}_{t, \boldsymbol{\epsilon}}\bigg[w(t)
    (\boldsymbol{\epsilon}_\phi(\mathbf{x}_t;y,t)-\boldsymbol{\epsilon})\dfrac{\partial \mathbf{x}_t}{\partial\mathbf{x}}\dfrac{\partial \mathbf{x}}{\partial{\theta}}\bigg],
\end{equation}
where $w(t)$ is a weighting function that depends on the timestep $t$ and $y$ denotes the given condition. $\epsilon_{\theta}(\mathbf{x}_t, t)$ is autoencoder in diffusion model to estimate the origin distribution of the given noise $\mathbf{x}_t$. 

\subsection{Physical Properties with Viscoelasticity Deformation}
\label{subsec:physcial_properties}
To model physical properties, we employ MLS-MPM~\cite{hu2018moving} as our simulator, and we formalize the simulation process for a single sub-step as follows:

\begin{equation}
  \mathbf{x^{n+1}}, \mathbf{v^{n+1}}, \mathbf{F^{n+1}}, C^{n+1} = \mathcal{S}(\mathbf{x^{n}}, \mathbf{v^{n}}, \mathbf{F^{n}}, C^{n}, {\theta}, \Delta t),
\end{equation}

Here, $\theta$ contains the physical properties of all particles: mass $m_i$, Young's modulus $E_i$, Poisson's ratio $\nu_i$, Lamé coefficients $\lambda, \mu$, viscosity coefficient $\nu_{Ni}$ and volume $V_i$. $\Delta t$ denotes the simulation step size. Within the MPM simulation, stress, as depicted in Equation~\ref{eq:decomposition}, can be divided into two components: one representing elastoplasticity, denoted as $\mathbf{F_E}$, and the other referring to viscoelasticity, denoted as $\mathbf{F_N}$. The two parts are also illustrated in Figure~\ref{fig:pipeline}, consisting of a parallel combination: (a) an elastic spring with a frictional element for plasticity and; (b) another spring with a viscous dash-pot element which is assumed to capture the dissipation. Now, we elaborate on the computation for each component individually. 

\textbf{Model for elastoplastic part.}
We first explain how to compute the internal stress $\boldsymbol{\sigma_E}$ through $\mathbf{F_E}$, which is essential in updating kinematical variables like velocities $v$ through Eq.~\ref{EOM}, for the elastic branch. We will speak out the rule here but postpone explanations and intuitions in Appendix~\ref{append:decomposition}. As we demonstrate in Sec.~\ref{subsec:preliminary} and Appendix~\ref{append:continuum_mechanics}, one can compute the Cauchy stress tensor given the energy function. In this work, we choose the fixed corotated constitutive model for the elastic part, whose energy function is
\begin{equation}
    \psi(\mathbf{F_E}) = \psi(\boldsymbol{\Sigma_E}) = \mu_E \sum_{i} (\sigma_{E,i} -1)^2+\frac{\lambda_E}{2}(\text{det}(\mathbf{F_E})-1)^2,
\end{equation}
where $\boldsymbol{\Sigma_E}$ is the diagonal singular value matrix $\mathbf{F_E}=\mathbf{U}\boldsymbol{\Sigma_E}\mathbf{V^T}$. And $\sigma_{E,i}$ are singular values of $\mathbf{F_E}$. From this energy function, we can compute the Cauchy stress tensor as:
\begin{equation}
    \boldsymbol{\sigma_E}=\frac{2 \mu}{\text{det}(\mathbf{F_E})}\left(\mathbf{F_E}-\mathbf{R}\right) \mathbf{F_E^T}+\lambda_E({\mathrm{det}}(\mathbf{F_E})-1),
\end{equation}
where $\mathbf{R}=\mathbf{U}\mathbf{V^T}$. We now explain how to update the $\mathbf{F^n_E}$ with the velocity field $\mathbf{v^n}$ at $n$th step. For purely elastic case, this can be simply done via:
\begin{equation}
    \mathbf{F^{n+1}_E} = (\mathbf{I}+\Delta t \nabla \mathbf{v^n}) \mathbf{F^n},
\end{equation}
where $\Delta t$ is the length of the time segment in the MPM method. This formula represents the fact that the internal velocity field causes a change in strains.

\textbf{Model for viscoelastic part.}
We now explain the other branch of our model: the viscoelastic part. The total strain now consist of two parts $\mathbf{F}=\mathbf{F_N}\mathbf{F_V}$. There are two key features of this brunch. First, only the $\mathbf{F_N}$ contributes to the internal stress. Secondly, $\mathbf{F_V}$ only plays a role in the update rule of $\mathbf{F_N}$. For the first step, we know have a relation between internal stress $\boldsymbol{\sigma_N}$ (or equivalently Kirchoff tensor $\boldsymbol{\tau_N}=\text{det}(\mathbf{F_N})\boldsymbol{\sigma_N}$):
\begin{equation}
    \tau_N = 2\mu_N \epsilon_N +\lambda_N \text{tr}(\epsilon_N) \mathbf{I}
\end{equation}
where we denote $\tau_N$ as a vector of singular value of $\boldsymbol{\tau_N}$, in another word $\boldsymbol{\tau_N}= U \text{diag}(\tau_N) V^T$. And $\epsilon_N$, called log principle Kirchoff tensor, denotes a vector takes the diagonal element of $\log \Sigma_N$, where $\Sigma_N$ as the diagonal singular value matrix $\mathbf{F_N}=U \Sigma_N V^T$. We can again update kinemetical variables afterward.
The second step is we need to modify the update rule for $\mathbf{F_N}$ tensor. This boils down to a trial-and-correction procedure where we first update
\begin{equation}
    \mathbf{F^{n}_{N,\text{tr}}} = (\mathbf{I}+\Delta t \nabla \mathbf{v^n}) \mathbf{F^n_N},
\end{equation}
then we modify the trial strain tensor $\mathbf{F^{n}_{N,\text{tr}}}$ by
\begin{equation}
    \epsilon_N^{n+1} = A(\epsilon_{N,\text{tr}}^n-B\text{tr}(\epsilon_{N,\text{tr}}^n)\cdot \mathbf{1})
\end{equation}
where $\epsilon_N^{n+1}$ denotes the log principle Kirchoff tensor of $\mathbf{F^{n+1}_{N}}$. $A$ and $B$ are functions of viscocity parameters $\nu_d$ and $\nu_e$. In this work, we take $\nu_d=\nu_e=\nu_N$ for simplicity.
Please refer to Appendix~\ref{append:decomposition} for more detailed intuitions and explanations.

\subsection{Optimization with Video Diffusion Model}
\label{subsec:optimization}
Through iterations of the Material Point Method (MPM), we obtain a set of Gaussians evolving over time $t$. To ensure consistency, the camera remains stationary as we capture and render each Gaussian to produce a reference image $I_t^r$. To supervise and optimize this process, we employ two optional models: image-to-video and text-to-video diffusion models. For video diffusion model, we utilize Stable Video Diffusion~\cite{svdblattmann2023} that learns rich 2D diffusion prior from large real-world data. Then, we use SDS loss to guide our optimization process as:
\begin{equation}
  \nabla_{\theta} \mathcal{L}_\text{SDS} = \mathbb{E}_{t, p, \mathbf{\epsilon}}
  \left[
  w(t)
  (\epsilon_\phi(I^p_t; t, I^r_t, \Delta p, y) - \epsilon) 
  \frac {\partial I^p_t} {\partial {\theta}}
  \right].
\end{equation}

Here, $w(t)$ represents a time-dependent weighting function, $\epsilon_\phi(\cdot)$ denotes the predicted noise generated by the 2D diffusion prior $\phi$, $\Delta p$ signifies the relative change in camera pose from the reference camera $r$, and $y$ denotes the given condition (\ie image or text). Furthermore, we adopt a partial filling strategy like~\cite{physgaussianxie2023}, where internal volumes of select solid objects are optionally filled to augment simulation realisticity. For better rendering quality, we optimize the filled Gaussian along with the SDS process. The optimization objective for filled Gaussians is defined as:

\begin{equation}
\label{eq:ref}
  \mathcal{L}_\text{Fill} = \frac 1 {T} \sum_{t=1}^T \lambda||I_t^p - I_t^r||^2_2
\end{equation}

The two losses mentioned above are optimized separately.

%% file: nips2024/4-experiments.tex
\section{Experiments}
\label{sec: experiments}

\begin{figure*}[!t]
    \centering
    \includegraphics[width=\linewidth]{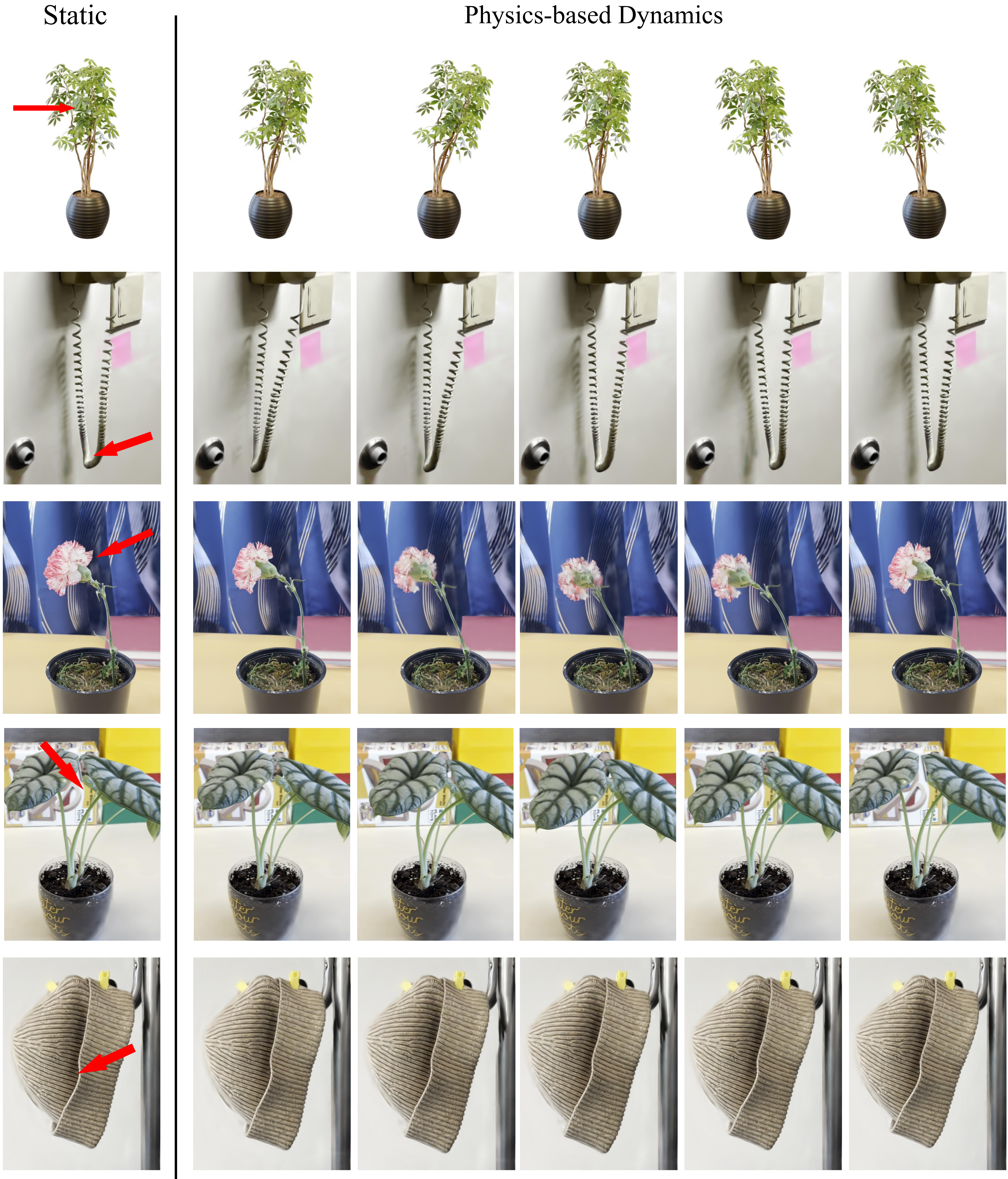}
    \caption{\textbf{Visual results of Physics3D} on different subjects with an external force (red arrow). Physics3D is able to generate realistic scene movement while maintaining good motion consistency.}
    \label{fig:visual_results}
    \vspace{-0.4cm}
\end{figure*}

In this section, we conduct extensive experiments to evaluate our Physics3D and show the comparison results against other methods~\cite {physgaussianxie2023, zhang2024physdreamer, ren2023dreamgaussian4d}. We first present our qualitative results and comparisons with baselines (Sec.~\ref{subsec:qualitative_results}). Then we report the quantitative results with a user study (Sec.~\ref{subsec: quantitative result}). Finally, we carry out more open settings and ablation studies to further verify the efficacy of our framework design (Sec.~\ref{subsec: ablation}). Please refer to the Appendix for more visualizations and detailed analysis.

\subsection{Experiment Setup}\label{subsec:experiment_setup}
\noindent \textbf{Datasets.}
We evaluate our method for generating diverse dynamics using several sources of input. We choose four real-world static scenes from PhysDreamer~\cite{zhang2024physdreamer} for fair comparison. Each scene includes an object and a background. The objects include a carnation, an alocasia plant, a telephone cord, and a beanie hat. We also employed NeRF datasets~\cite{mildenhall2021nerf-eccv} to evaluate the efficacy of our proposed model. Additionally, we utilize BlenderNeRF~\cite{Raafat_BlenderNeRF_2023} to synthesize several scenes.

\noindent \textbf{Implementation Details.} In our implementation, we initiate the process by reconstructing 3D Gaussians from multi-view images, establishing a foundational representation of the scene. In complex realistic cases, we undertake a segmentation step to differentiate between the background and foreground elements, focusing solely on the latter for subsequent simulation tasks. Prior to simulation, we execute internal particle filling operations to refine the representation further. Each Gaussian kernel is then associated with a distinct set of physical properties targeted for optimization, facilitating the fine-tuning process. Subsequently, we discretize the foreground region into a grid structure, typically set at dimensions of $50^3$. As for the MPM simulation, we employ 400 sub-steps within each temporal interval spanning successive video frames. This temporal granularity translates to a sub-step duration of $1e-4$ second, ensuring precision and accuracy in the simulation dynamics. Notably, the optimization process for each object requires approximately 5 minutes to complete on a single NVIDIA A6000 (48GB) GPU.

\noindent \textbf{Baselines and Metrics.} We extensively compare our method with three baselines: PhysDreamer~\cite{zhang2024physdreamer}, PhysGaussian~\cite{physgaussianxie2023} and DreamGaussian4D~\cite{ren2023dreamgaussian4d}. Following~\cite{zhang2024physdreamer}, we show our results with notable comparisons with space-time slices. For metrics, we evaluate our approach with the video-quality metrics: PSNR, SSIM, MS-SSIM, and VMAF on the rendered videos. We also conduct a user study to further demonstrate the overall quality and the alignment with real-world physics in our simulation, which can be found in our Appendix.

\subsection{Qualitative Results}
\label{subsec:qualitative_results}
Figure~\ref{fig:visual_results} shows qualitative results of simulated interactive motion. For each case, we visualize one example with its initial scene and deformation sequence. The results demonstrate our model's capability of simulating the movement of complex textured objects, presenting a realistic and physically plausible outcome. Additional experiments are included in the Appendix~\ref{append:visual_results}.

\begin{figure*}[!t]
    \centering
    \includegraphics[width=\linewidth]{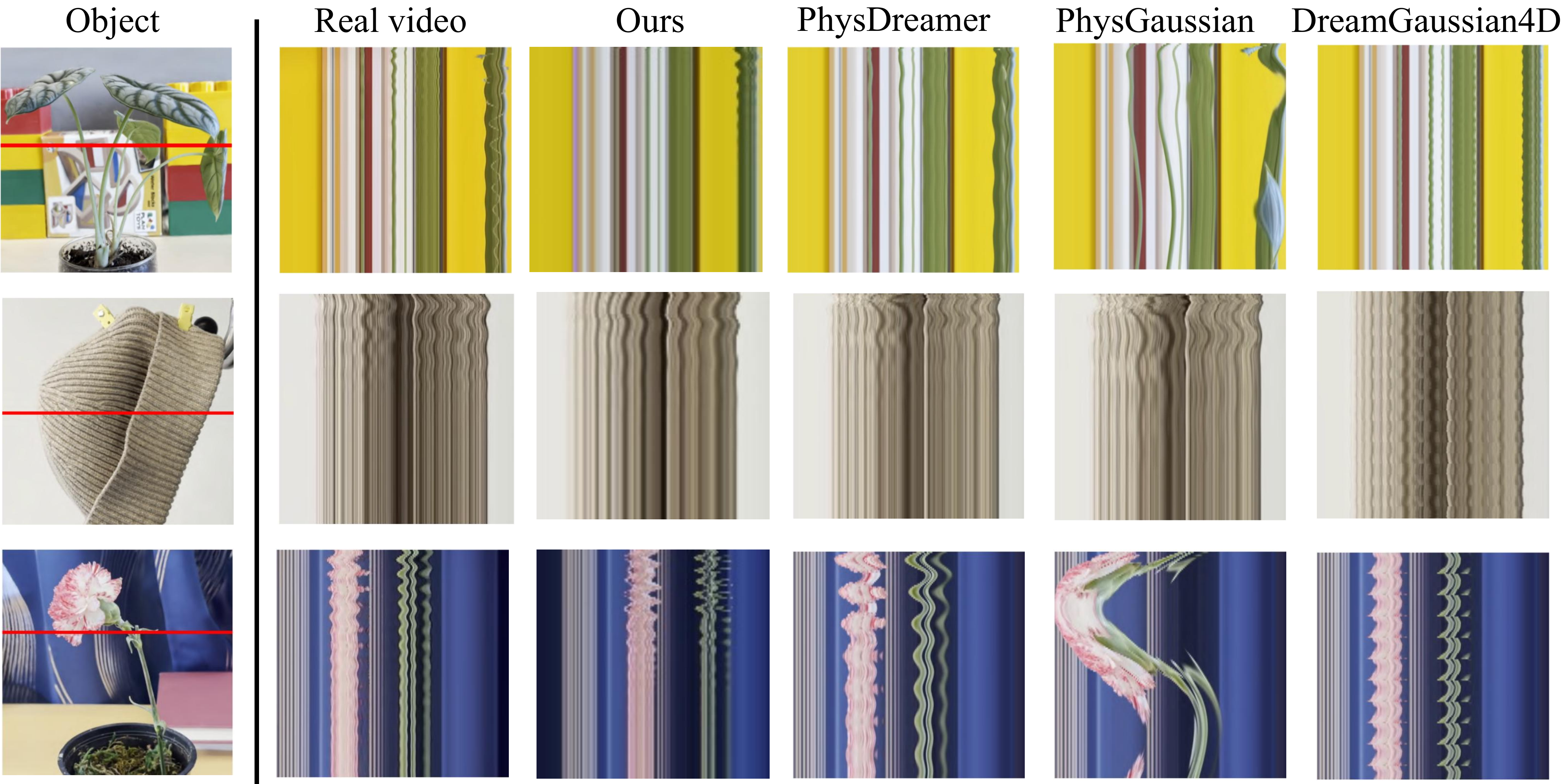}
    \vspace{-0.6cm}
    \caption{\textbf{Comparison results} on different subjects. From the visual results, we observe that PhysDreamer~\cite{zhang2024physdreamer} only estimates the elastic properties of objects, resulting in the lack of damping. DreamGaussian4D~\cite{ren2023dreamgaussian4d} and PhysGaussian~\cite{physgaussianxie2023} are respectively limited in unrealistically constant, low-magnitude periodic motion and low-frequency movements. In contrast, our model successfully balances high and low-frequency oscillations with more realistic damping.}
    \label{fig:comparison}
    \vspace{-0.4cm}
\end{figure*}
Following PhysDreamer~\cite{zhang2024physdreamer}, we compare our results with real captured videos and simulations from other methods~\cite{zhang2024physdreamer, physgaussianxie2023, ren2023dreamgaussian4d} in Figure~\ref{fig:comparison}. We utilize space-time slices to present our comparisons. These slices depict time along the vertical axis and spatial slices of the object along the horizontal axis, as indicated by the red lines in the "object" column. Through these visualizations, we aim to elucidate the magnitude and frequencies of the oscillating motions under scrutiny.
PhysDreamer~\cite{zhang2024physdreamer} closely approximates the elastic properties of objects, resulting in periodic oscillations with subtle damping, contrasting the unrealistic aspects. PhysGaussian~\cite{physgaussianxie2023} showcases unrealistically low-frequency movements due to inaccurate parameter settings. DreamGaussian4D~\cite{ren2023dreamgaussian4d} lacks physical prior and generates unrealistically constant, low-magnitude periodic motion. In contrast, our model successfully balances high and low-frequency oscillations with more realistic damping, aligning more closely with the behavior of objects in the real world.

\subsection{Quantitative Results}
\label{subsec: quantitative result}
\begin{wraptable}{r}{8.35cm}
    \centering
    \small
    \vspace{-7mm}
    \caption{\textbf{Quantitative comparisons} on rendered videos using different video-quality metrics.
    }
    \label{tab:precision_metrics}
    \vspace{1mm}
    \resizebox{0.6\textwidth}{!}{
    \begin{tabular}{lcccc}
        \toprule
        & PSNR$\uparrow$ & SSIM$\uparrow$ & MS-SSIM$\uparrow$ & VMAF$\uparrow$ \\
        \midrule
        PhysDreamer~\cite{zhang2024physdreamer} & 13.89 & 0.55 & 0.37 & 0.52\\
        PhysGaussian~\cite{physgaussianxie2023} & 13.86 & 0.57 & 0.39 & 0.59 \\
        Ours & \textbf{14.72} & \textbf{0.59} & \textbf{0.49} & \textbf{0.59}\\
        - \textit{w/o} viscoelastic part & 14.13 & 0.53 & 0.36 & 0.52 \\
        - \textit{w/o} elastoplastic part & 13.53 & 0.55 & 0.41 & 0.50 \\ 
        \bottomrule
    \end{tabular}}
    \vspace{-5mm}
\end{wraptable}
Table~\ref{tab:precision_metrics} shows the average video-quality metrics over rendered videos from the same fixed perspective. Results clearly demonstrate higher scores for Physics3D, indicating better video quality and motion consistency of our results. We also conduct numerous user experiments in Appendix~\ref{append:user_study}.

\subsection{Ablation Study}
\label{subsec: ablation}
\begin{figure*}[ht]
    \centering
    \includegraphics[width=\linewidth]{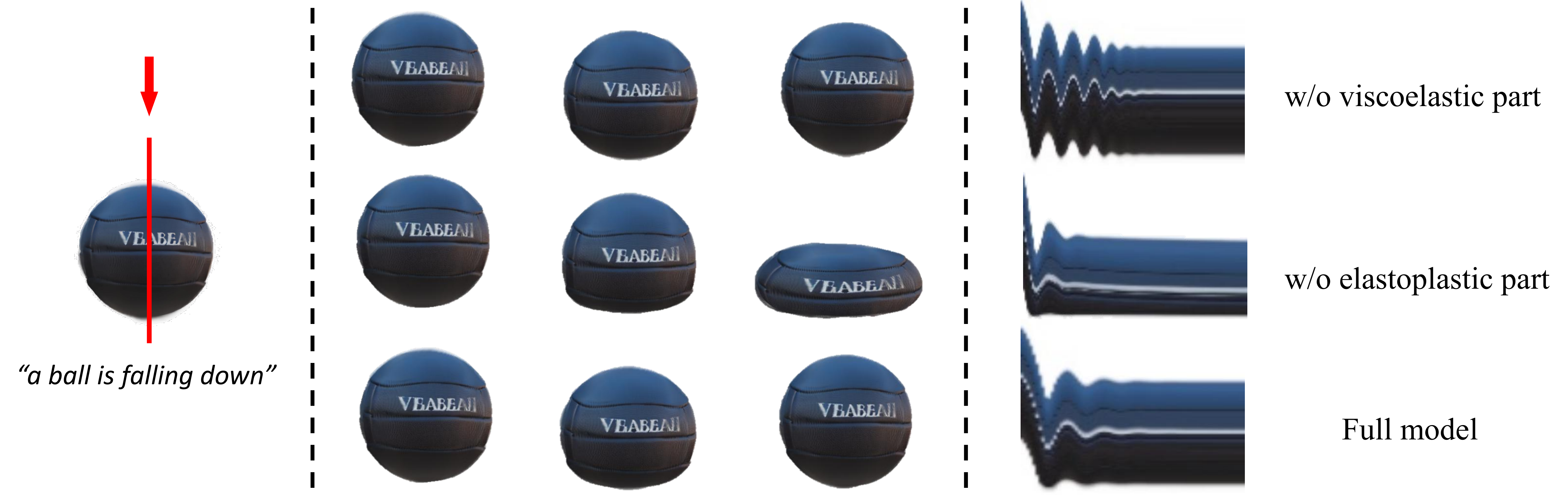}
    \vspace{-0.5cm}
    \caption{Ablation study on elastoplastic and viscoelastic parts of our model.}
    \label{fig:ablation}
    \vspace{-0.4cm}
\end{figure*}

We conduct ablation study in Figure~\ref{fig:ablation} to evaluate the efficacy of our physics modeling process. Specifically, we investigate the importance of elastoplastic and viscoelastic components from the model architecture. We observe that the removal of either of these modules leads to a notable degradation in the realism of the physical simulations. Particularly, the absence of the elastoplastic component results in a lack of elasticity, making objects more susceptible to shape deformation and fluid-like behavior. On the other hand, the absence of the viscoelastic component leads to a deficiency in sustained damping and rebound effects, especially in scenarios with minimal external disturbances where energy dissipation occurs rapidly. These show the significance of both components of our model in capturing the dynamics of physical objects. Please refer to our Appendix for more ablations.

%% file: nips2024/5-conclusion.tex
\section{Conclusion}
In this paper, we present Physics3D, a novel framework to learn various physical properties of 3D objects from video diffusion model. Our method tackles the challenge of estimating diverse material properties by incorporating two key components: elastoplastic and viscoelastic modules. The elastoplastic component facilitates simulations of pure elasticity, while the viscoelastic module introduces damping effects, crucial for capturing the behavior of materials exhibiting both elasticity and viscosity. Furthermore, we leverage a video generation model to distill inherent physical priors, improving our understanding of realistic material characteristics. Extensive experiments show Physics3D is effective in creating high-fidelity and realistic 3D dynamics.

\label{subsec:limitation}
\noindent \textbf{Limitations and Future Work.} 
In complex environments with a lot of entangled objects, our method requires manual intervention to assign the scope of movable objects and define the filling ranges for objects, which is not efficient for more real applications. In the future, we aim to utilize the prior of large segmentation models to solve the problem with more comprehensive physics system modeling. We believe that Physics3D takes a significant step to open up a wide range of applications from realistic simulations to interactive virtual experiences and will inspire more works in the future.

%% file: nips2024/appendix.tex
\section*{Appendix}
\section{More Results}
To further demonstrate the effectiveness and impressive visualization results of our Physics3D, we conducted more experiments including additional visual results and user study.

\subsection{Additional Visual Results}
\label{append:visual_results}
\begin{figure*}[ht]
    \centering
    \includegraphics[width=\linewidth]{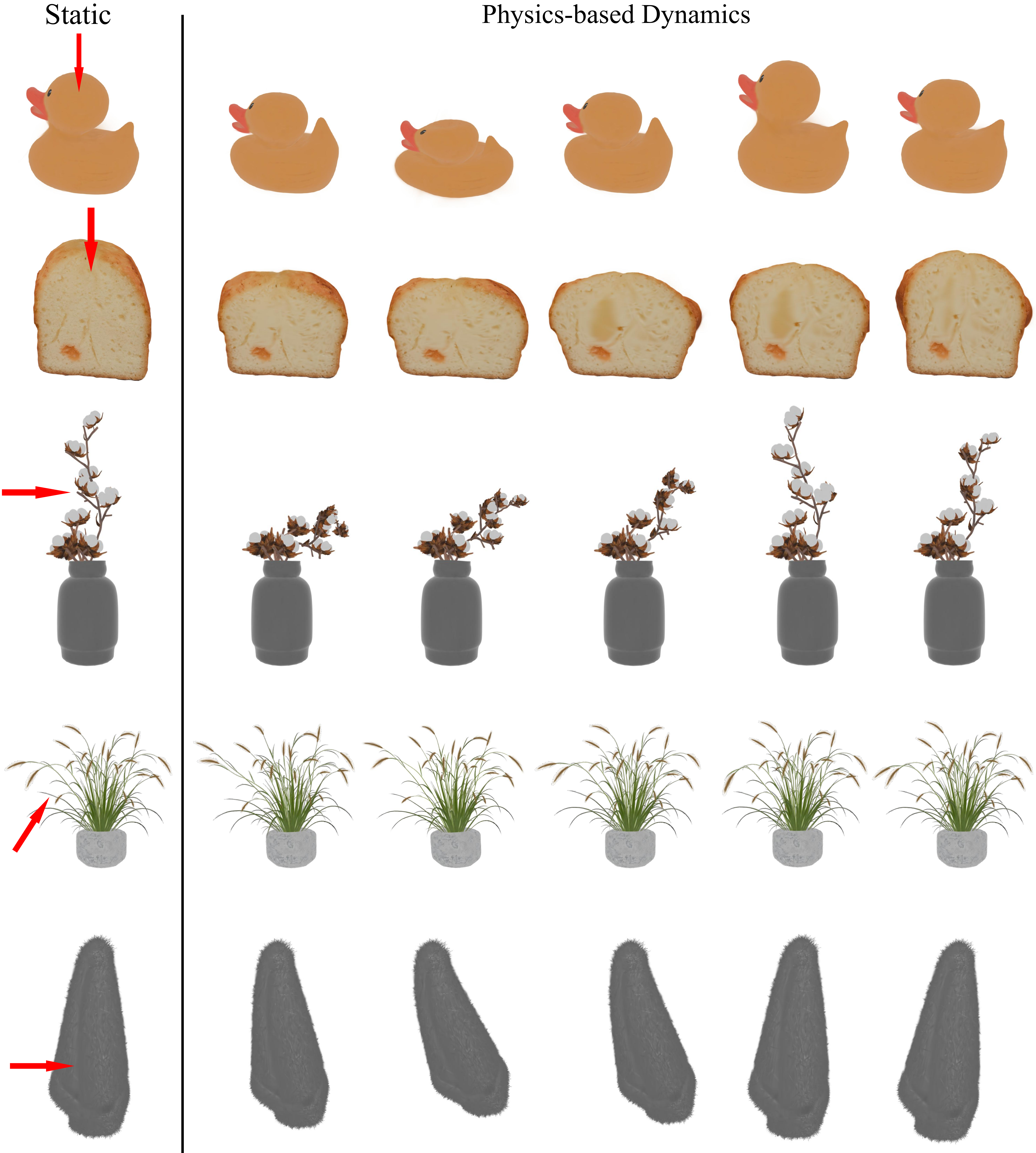}
    \caption{More visual results on different objects. Notice ours can simulate a variety of non-elastic and composite materials, including \textit{rubber}, \textit{fluffy bread}, \textit{fabric} etc.}
    \label{fig:more_visual_results}
\end{figure*}

Figure~\ref{fig:more_visual_results} shows additional visual results of interactive video sequences. We use BlenderNeRF~\cite{Raafat_BlenderNeRF_2023} to synthesize the cases below. For each case, We apply external forces of different directions and magnitudes to static objects and render video frames to show their motion states.

\subsection{Additional Ablation Study}\label{append:ablation}
We carry out additional ablation analyses on the Physics3D design in Figure~\ref{fig:ablation} to assess the efficacy of our physics modeling process. Specifically, we investigate the impact of removing the elastoplastic and viscoelastic components from the model architecture. The findings highlight that the removal of either of these modules leads to a notable decrease in the realisticity of the physical simulations. Particularly, the absence of the elastoplastic component results in a lack of elasticity, thereby making objects more susceptible to shape deformation and fluid-like behavior. On the other hand, the absence of the viscoelastic component leads to a deficiency in sustained damping and rebound effects, especially in scenarios with minimal external disturbances where energy dissipation occurs rapidly. These outcomes underscore the significance of both components of our model in capturing the dynamics of physical objects.
Notice that without internal filling, the hollow part of the cake is prone to collapse in the simulation process, but with internal filling, the object's deformation under external force is more realistic and smooth.

\begin{figure*}[ht]
    \centering
    \includegraphics[width=\linewidth]{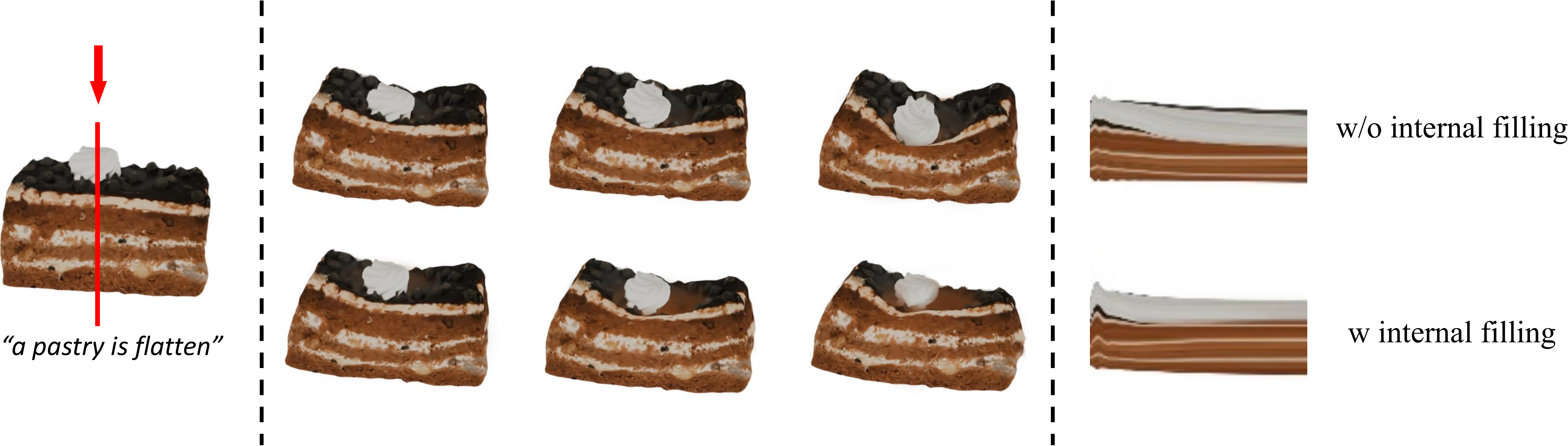}
    \caption{Ablation study on internal filling.}
    \label{fig:ablation_filling}
    \vspace{-0.4cm}
\end{figure*}

\subsection{User Study}\label{append:user_study}
For user study, we show each volunteer with five samples of rendered video from a random method (PhysDreamer~\cite{zhang2024physdreamer}, PhysGaussian~\cite{physgaussianxie2023}, DreamGaussian4D~\cite{ren2023dreamgaussian4d}, and ours) and real-captured videos. The study engaged 30 volunteers to assess the generated results in 20 rounds. In each round, they were asked to select the video they preferred the most, based on quality, realisticity, alignment with input 3D object, and fluency. We find our method is significantly preferred by users over these aspects. 

\begin{table}[h]
    \centering
    \small
    \caption{\textbf{Quantitative comparison results} of PhysGaussian~\cite{physgaussianxie2023}, PhysDreamer~\cite{zhang2024physdreamer} and our Physics3D on the action coherence, motion realism, and overall quality score in a user study, rated on a range of 1-10, with higher scores indicating better performance.}
    \label{tab:user_study}
    \vspace{1.5mm}
    \begin{tabular}{lccc}
        \toprule
        {Method}  & Action Coherence & Motion Realism & Overall Quality \\
        \midrule
        PhysGaussian~\cite{physgaussianxie2023} & 7.82 & 6.89 & 6.93 \\
        PhysDreamer~\cite{zhang2024physdreamer} & 8.76 & 7.73 & 7.89 \\
        Physics3D (Ours) & \textbf{8.95} &  \textbf{8.57}  & \textbf{9.05}     \\     
        \bottomrule
    \end{tabular}
\end{table}

\section{More Analysis}
\subsection{Material Point Method (MPM) Algorithm}

The Material Point Method (MPM) simulates the behavior of materials by discretizing a continuum body into particles and updating their properties over time. Here's an additional summary of the MPM algorithm:

\textbf{Particle to Grid Transfer.} Mass and momentum are transferred from particles to grid nodes. This step involves distributing the particle properties (mass and velocity) to nearby grid points. 
\begin{align*}
    m_i^n &= \sum_p w_{ip}^n m_p, \\
    m_i^n \mathbf{v}_i^n &= \sum_p w_{ip}^n m_p(\mathbf{v_p^n} + \mathbf{C}_p^n(\mathbf{x_i} - \mathbf{x_p^n})).
\end{align*}

\textbf{Grid Update.} Grid velocities are updated based on external forces and the forces from neighboring particles. This step moves the grid points according to the applied forces.
\begin{equation*}
    \mathbf{v_i^{n+1}} = \mathbf{v_i^{n}} - \frac{\Delta t}{m_i} \sum_p \boldsymbol{\tau}_p^n \nabla w_{ip}^n V_p^0 + \Delta t \mathbf{g}.
\end{equation*}

\textbf{Grid to Particle Transfer.} Velocities are transferred back to particles, and particle states are updated. This step brings the changes in grid velocities back to the particles.
\begin{align*}
    \mathbf{v_p^{n+1}} &= \sum_i \mathbf{v_i^{n+1}} w_{ip}^n, \\
    \mathbf{x_p^{n+1}} &= \mathbf{x_p^{n}} + \Delta t \mathbf{v_p^{n+1}}, \\
    {\boldsymbol{C}}_p^{n+1}&=\frac{12}{\Delta x^2(b+1)} \sum_{{i}} w_{{i} p}^n \mathbf{{v}_i^{n+1}} \left(\mathbf{{x}_{{i}}^n}-\mathbf{{x}_p^n}\right)^T,\\
    \nabla{\mathbf{v_p^{n+1}}} &= \sum_i \boldsymbol{v}_{{i}}^{n+1} {\nabla{w_{{i} p}^n}^{T}},\\
    \tau_p^{n+1} &= \boldsymbol{\tau}(\mathbf{F_E^{n+1}}, \mathbf{F_N^{n+1}}), \\
\end{align*}
Here $b$ is the B-spline degree, and $\Delta x$ is the Eulerian grid spacing. We extensively demonstrate how to update $\mathbf{F_E}$, $\mathbf{F_N}$ and $\tau$ in Appendix~\ref{append:continuum_mechanics}.

\subsection{Dynamic 3D Generation.}
Dynamic 3D generation aims to synthesize the dynamic behavior of 3D objects or scenes in the process of generating three-dimensional representations. Unlike static 3D generation methods~\cite{liu2023sherpa3d, hong2023lrm, wang2024crm, liu2024make} that solely focus on the spatial morphology of objects, the field of 3D dynamics imposes a higher requirement, necessitating the incorporation of information from all three spatial dimensions as well as the temporal dimension. As advancements in 3D representation techniques continue to emerge, certain parameterizable 3D representations have empowered us to inject dynamic information into 3D models~\cite{li2022pac, kratimenos2023dynmf}. One popular approach~\cite{newcombe2015dynamicfusion, ren2023dreamgaussian4d} integrates three essential components:

\textbf{Gaussian Splatting}~\cite{kerbl20233d-gaussian} represents 3D information with a set of 3D Gaussian kernels. Each Gaussian can be described with a center $x \in \mathbb{R}^3$, a scaling factor $s \in \mathbb{R}^3$, and a rotation quaternion $q \in \mathbb{R}^4$. Additionally, an opacity value $\alpha \in \mathbb{R}$ and a color feature $c \in \mathbb{R}^3$ are for volumetric rendering. These parameters can be collectively denoted by $\theta$, with $\theta_i = \{x_i, s_i, q_i, \alpha_i, c_i\}$ representing the parameters for the $i$-th Gaussian. The volume rendering color $C$ of each pixel is computed by blending $N$ ordered points overlapping the pixel:
\begin{equation}
    C = \sum_{i \in N} c_i \alpha_i  \prod_{j=1}^{i-1}(1-\alpha_j).
\end{equation}

\label{append:diffusion}
\textbf{Diffusion Model}~\cite{ddpm, improved_ddpm}
is usually pre-trained on large 2D datasets to provide a foundational motion prior for dynamic 3D generation. These models, characterized by their probabilistic nature, are tailored to acquire knowledge of the data distribution $p(x_0)$ through a step-by-step denoising process applied to a normally distributed variable. Throughout the training phase, the data distribution is perturbed towards an isotropic Gaussian distribution over $T$ timesteps, guided by a predefined noising schedule $\alpha_t \in (0,1) $, where $\overline{\alpha_t} = \sum_{s=1}^t \alpha_s $ and $t$ uniformly sampled from ${1, . . . , T }$:
\begin{equation}
    \mathbf{z}_{t} =\sqrt{\bar{\alpha}_t} \mathbf{x_0} + \sqrt{1-\bar{\alpha}_{t}}\boldsymbol{\epsilon}, \text{ where } \boldsymbol{\epsilon} \sim \mathcal{N}(\mathbf{0}, \mathbf{I}).
\end{equation}
The backward denoising process estimates the origin distribution by autoencoder $\epsilon_{\theta}(\mathbf{z}_t, t)$. The final training loss can be simplified as:
\begin{equation}\label{eq:diffusion}
    \mathcal{L}_{DM} = \mathbb{E}_{\mathbf{x}, \boldsymbol{\epsilon} \sim \mathcal{N}(\mathbf{0},\mathbf{I}) }\left[\left\|\boldsymbol{\epsilon} - \boldsymbol{\epsilon}_{\phi}\left(\mathbf{z}_t, t\right) \right\|^{2}_{2}\right].
\end{equation}

\textbf{Score Distillation Sampling} (SDS) is a technique used in machine learning, particularly in the context of generative models, to improve sample quality and diversity. It addresses the challenge of generating high-quality samples from complex probability distributions, such as those learned by deep neural networks.

In traditional generative models like Generative Adversarial Networks (GANs)~\cite{goodfellow2020generative} or Variational Autoencoders (VAEs)~\cite{kingma2013auto}, generating samples involves directly sampling from the learned latent space. However, this approach often results in low-quality samples with poor diversity, especially in regions of low probability density.

SDS introduces a novel sampling strategy that leverages the notion of score matching. Score matching, introduced by Hyv{\"a}rinen in 2005~\cite{hyvarinen2005estimation}, is a technique for training generative models by matching the score function (gradient of the log-density) of the model distribution to that of the true data distribution.

Given a target distribution $p_{\text{data}}(\mathbf{x})$ and a model distribution $p_{\phi}(\mathbf{x})$, where $\phi$ represents the parameters of the model, the score function is defined as:
\begin{equation}
\nabla_{\mathbf{x}} \log p_{\phi}(\mathbf{x})
\end{equation}

The score function provides valuable information about the local geometry of the probability distribution, helping to guide the sampling process towards regions of high probability density.

In SDS, the goal is to distill the score function learned by a complex generative model into a simpler, more tractable form. This distilled score function can then be used to guide the sampling process efficiently, resulting in higher-quality samples with improved diversity.

Formally, given a set of generated samples $\{\mathbf{x}_i\}_{i=1}^{N}$ from the model distribution $p_{\phi}(\mathbf{x})$, the distilled score function $\hat{s}(\mathbf{x})$ is learned to approximate the score function of the model distribution. This is achieved by minimizing the score matching loss:
\begin{equation}
\mathcal{L}_{\text{SM}}(\hat{s}) = \frac{1}{N}\sum_{i=1}^{N} \|\nabla_{\mathbf{x}} \log p_{\theta}(\mathbf{x}_i) - \hat{s}(\mathbf{x}_i)\|^2
\end{equation}

where $\|\cdot\|$ denotes some norm (e.g., $L^2$ norm) and $\hat{s}(\mathbf{x}_i)$ is the estimated score at sample $\mathbf{x}_i$. 

As for particular diffusion model, its score function can be related to the predicted noise (shown in Eq.~\ref{eq:diffusion})  for the smoothed density through Tweedie’s formula~\cite{Robbins1992}:
\begin{equation}
\boldsymbol{\epsilon}_{\phi}\left(\mathbf{z}_t, t\right) = -\sigma_ts_\phi(\mathbf{z}_t;t).
\end{equation}

Training the diffusion model with a (weighted) evidence lower bound (ELBO) simplifies to a weighted denoising score matching objective for parameters $\phi$ \citep{ddpm,kingma2021on}:
\begin{equation}
\mathcal{L}_{\text{Diff}}(\phi, \mathbf{x}) = \mathbb{E}_{t\sim \mathcal{U}(0,1), \epsilon \sim \mathcal{N}(\mathbf{0}, \mathbf{I})} \left[w(t)\|\epsilon_\phi(\alpha_t \mathbf{x} + \sigma_t \epsilon; t) - \epsilon\|^2_2\right] \, ,
\label{eq:train}
\end{equation}
where $w(t)$ is a weighting function that depends on the timestep $t$. To understand the difficulties of this approach, consider the gradient of $\mathcal{L}_{\text{Diff}}$:
\begin{align}
\nabla_\theta \mathcal{L}_{\text{Diff}}(\phi, \mathbf{x}=g(\theta)) &= \mathbb{E}_{t, \epsilon}\Bigg[w(t)\underbrace{\left(\hat\epsilon_\phi(\mathbf{z}_t; y, t)  - \epsilon\right) \vphantom{{\frac{\partial \hat\epsilon_\phi(\mathbf{z}_t; y, t)}{\partial \mathbf{z}_t}}}}_{\text{Noise Residual}} \underbrace{{\frac{\partial \hat\epsilon_\phi(\mathbf{z}_t; y, t)}{\partial \mathbf{z}_t}}}_{\text{U-Net Jacobian}} \underbrace{\vphantom{{\frac{\partial \hat\epsilon_\phi(\mathbf{z}_t; y, t)}{\partial \mathbf{z}_t}}}\frac{\partial \mathbf{x}}{\partial \theta}}_{\text{Generator Jacobian}}\Bigg]
\label{eq:graddiff}
\end{align}

where following DreamFusion~\cite{poole2022dreamfusion}, we absorb the constant $\alpha_t {\bf I} = \partial \mathbf{z}_t / \partial x$ into $w(t)$. In practice, the U-Net Jacobian term is expensive to compute (requires backpropagating through the diffusion model U-Net), and poorly conditioned for small noise levels as it is trained to approximate the scaled Hessian of the marginal density. 
In~\cite{poole2022dreamfusion}, It is found that omitting the U-Net Jacobian term leads to an effective gradient:
\begin{align}
\nabla_{\theta} \mathcal{L}_{\text{SDS}}(\phi, \mathbf{x}=g(\theta)) \triangleq \mathbb{E}_{t, \epsilon}\left[w(t)\left(\hat\epsilon_\phi(\mathbf{z}_t; y, t)  - \epsilon\right) \frac{\partial \mathbf{x}}{\partial \theta}\right],
\label{eq:sdsgrad}
\end{align}
here, we get the gradient of a weighted probability density distillation loss in Eq.~\ref{eq:sds}.

\label{append:continuum_mechanics}
\subsection{Additional Analysis in Continuum Mechanics}
\textbf{Basic intuitions in continuum mechanics} involves explaining the strain tensor $\mathbf{F}$ which describes the internal deformation of the material, and $\boldsymbol{\sigma}$ which describe the internal stress tensor. In an actual material, there might be multiple compounds, each of them might have their own strain and stress tensor and they might interact with each other. For purpose of this subsection, we explain the intuition when there is only one such compounds. However we will generalize it into multiple compounds case later in the paper.

One could describe an equilibrium material as a point cloud $\mathbf{X}$. In actually simulation, one take a discretization of the system. For our convenience, we will just state the intuition in continum, and generalization to discrete points should be straightforward.  When the material is away from its equilibrium position, the position of point is given by $\mathbf{x} \neq \mathbf{X}$. However, we notice that if, for example, we move $\mathbf{x}$ and the adjacent point $\mathbf{\tilde{x}}$ in the same way, then there is actually no internal deformation of the material. In another word, internal deformation describe how relative position between two points $\mathbf{x}$ and its adjacent $\mathbf{\tilde{x}}$ change. To make this quantity well-defined, we normalize it with the original difference between $\mathbf{X}$ and $\mathbf{\tilde{X}}$
\begin{equation}
    \mathbf{\tilde{F}}_{ij} = \frac{\mathbf{\tilde{x}_i}-\mathbf{x}_i}{\mathbf{\tilde{X}_j}-\mathbf{X}_j} 
\end{equation}
Taking continuum limit, the equation becomes differential and reproduce
\begin{equation}
    \mathbf{F} = \frac{\partial \phi}{\partial \mathbf{X}} (\mathbf{X}, t)
\end{equation}
in the main text. 

To get the dynamics, we will further need the stress tensor $\boldsymbol{\sigma}$. However, the general relation between $\boldsymbol{\sigma}$ and $\mathbf{F}$ is usually complicated. In fact only when internal stress is a conservative force, we can related it to $\mathbf{F}$ through an energy function $\psi(\mathbf{F})$. An analog for this case is the Hooke's force in a spring, where we can introduce a potential energy for the force. There are two main types of internal stress people use in the literature. The first Piola-Kirchoff stress
\begin{equation}
   \mathbf{P} = \frac{\partial \psi}{\partial \mathbf{F}}
\end{equation}
and a related Cauchy stress
\begin{equation}
    \boldsymbol{\sigma} = \frac{1}{{\mathrm{det}}(\mathbf{F})} \frac{\partial \psi}{\partial \mathbf{F}} \mathbf{F}^T
\end{equation}
We will use Cauchy stress for most part of out paper. However, we could sometimes use the other one interchangeably. Physically, if we want the force acting on a unit surface with normal vector $\hat{n}$, the force will be $\boldsymbol{\sigma}\cdot \hat{n}$ . Another version people use is the Kirchoff stress $\boldsymbol{\tau} = \text{det}(\mathbf{F})\boldsymbol{\sigma}$. The only difference between them are whether it is measured with the deformed volume($\boldsymbol{\sigma}$) or the undeformed volume($\boldsymbol{\tau}$). Sometimes using one or the other could be easier for technical reason, as we can see later in the viscoelastic model. 

Beyond this simple case, the relation will be quite complicated and non-universal. In this paper, we discuss a specific variant of them, which we explain below. One can see both the relation between $\boldsymbol{\sigma}$ and $\mathbf{F}$, and the update rule for $\mathbf{F}$ get modified.

With these background, we can explain the intuition behind Eq.\ref{EOM}. The second equation 
\begin{equation}
    \frac{D\rho}{Dt} + \rho \nabla \cdot \mathbf{v} = 0,
\end{equation}
is nothing but mass conservation, $\rho \nabla \cdot \mathbf{v}$ is the local density times the local divergence, which physically correspond to the mass flows out of a local unit volumn in unit time.

The first equation is Newton's law
\begin{equation}
    \rho \frac{D \mathbf{v} }{D t} = \nabla \cdot \boldsymbol{\sigma} + \mathbf{f},
\end{equation}
the $\mathbf{f}$ term is the external force. The term $\nabla \cdot \boldsymbol{\sigma}$ is the divergence for the stress tensor, thus giving the total force on a unit volume material.

\label{append:decomposition}
\textbf{Model for elastoplastic part.}
We explain how to compute the internal stress $\boldsymbol{\sigma_E}$ through $\mathbf{F_E}$, and how to update $\mathbf{F_E}$ for the elastic part in this appendix. Again for current paper we consider the elastoplastic branch with only elastic part. As we have reviewed, one can compute the Cauchy stress tensor given the energy function. In this work, we choose the Fixed corotated constitutive model for the elastic part, whose energy function is
\begin{equation}
    \psi(\mathbf{F_E}) = \psi(\boldsymbol{\Sigma_E}) = \mu \sum_{i} (\sigma_{E,i} -1)^2+\frac{\lambda}{2}(\text{det}(\mathbf{F_E})-1)^2
\end{equation}
in which $\boldsymbol{\Sigma_E}$ is the diagonal singular value matrix $\mathbf{F_E}=\mathbf{U}\boldsymbol{\Sigma_E}\mathbf{V^T}$. And $\sigma_{E,i}$s are singular values of $\mathbf{F_E}$. From this energy function, one can compute the Cauchy stress tensor
\begin{equation}
    \boldsymbol{\sigma_E}=\frac{2 \mu}{\text{det}(\mathbf{F_E})}\left(\mathbf{F_E}-\mathbf{R}\right) \mathbf{F_E^T}+\lambda({\mathrm{det}}(\mathbf{F_E})-1)
\end{equation}
in which $\mathbf{R}=\mathbf{U}\mathbf{V^T}$. 

We now explain how to update the $\mathbf{F^n_E}$ with the velocity field $\mathbf{v^n}$ at $n$th step. For purely elastic case, this can be simply done via
\begin{equation}
    \mathbf{F^{n+1}_E} = (\mathbf{I}+\Delta t \nabla \mathbf{v^n}) \mathbf{F^n}
\end{equation}
in which $\Delta t$ is the length of the time segment in the MPM method. This formula represents the fact that internal velocity field cause change in strains. However, as one will say for the viscoelastic part, this rule will change.

\textbf{Model for viscoelastic part.}
We now explain the other branch of our model: the viscoelastic part. Now we should imagine a elastic strain series connect with a viscous dissipator. In another word, the total strain $\mathbf{F}=\mathbf{F_N}\mathbf{F_V}$. We now explain two key features for this brunch. First, only the $\mathbf{F_N}$ contributes to the internal stress. Secondly, $\mathbf{F_V}$ only plays a role in the update rule of $\mathbf{F_N}$. Very roughly, one can view the dissipator $\mathbf{F_V}$ as a reservoir for strain and each time during the update, people will be able to relocate part of the strain into the $\mathbf{F_V}$, such that only part of the strain contributes to the internal stress. 

For the $\mathbf{F_N}$ part, we again take it to be a elastic system, so the relation between strain and stress is again given by the energy function. Here we choose the energy function following \cite{fang2019silly}
\begin{equation}
    \psi_N(\Sigma_N)=\mu_N \text{tr}((\log \Sigma_N)^2)+\frac{1}{2}\lambda_N (\text{tr}(\log \Sigma_N))^2
\end{equation}
In this formula, we introduce $\Sigma_N$ as the diagonal singular value matrix $\mathbf{F_N}=U \Sigma_N V^T$. As one can see, this potential is only a function of the singular values, so the Kirchoff tensor $\boldsymbol{\tau_N}$ will be diagonal in the same basis as the strain $\mathbf{F_N}$. So we will just write down the relation between $\mathbf{F_N}$ and $\boldsymbol{\tau_N}$ just in the singular value basis
\begin{equation}
    \tau_N =\frac{\partial \psi_N(\Sigma_N)}{\partial \epsilon_N} = 2\mu_N \epsilon_N +\lambda_N \text{tr}(\epsilon_N) \mathbf{I}
\end{equation}
where we denote $\tau_N$ as a vector of singular value of $\boldsymbol{\tau_N}$, in another word $\boldsymbol{\tau_N}= U \text{diag}(\tau_N) V^T$. This is sometimes called principle Kirchoff tensor in literature. And $\epsilon_N$, called log principle Kirchoff tensor, denotes a vector takes the diagonal element of $\log \Sigma_N$. So this relation is essentially a vector equation. We can recover the Kirchoff tensor $\boldsymbol{\tau_N}$ thus the Cauchy stress tensor
\begin{equation}
    \boldsymbol{\sigma_N}=\frac{1}{\det(\mathbf{F_N})} \boldsymbol{\tau_N}.
\end{equation} 

Now we can discuss the update rule for the strain tensor $\mathbf{F_N}$. As we explained before, the rough intuition was part of the strain tensor $\mathbf{F_N}$ can dissipate into the dissipator $\mathbf{F_V}$. More quantitatively, we can follow the dissipator model in \cite{fang2019silly}, which results into a trial-and-correction procedure. The idea is we first imagine $\mathbf{F_N}$ evolve elastically
\begin{equation}
    \mathbf{F^{n}_{N,\text{tr}}} = (\mathbf{I}+\Delta t \nabla \mathbf{v^n}) \mathbf{F^n_N}
\end{equation}
Next step, we modify this trial strain tensor by introducing a dissipation into the dissipator
\begin{equation}\label{append:viscupdate}
    \epsilon_N^{n+1} = \epsilon_{N,\text{tr}}^n -\Delta t \frac{\partial \psi_V}{\partial \tau_N} 
\end{equation}
Here again, we assume the dissipation happen to only the singular value of the strain tensor, so $\epsilon_{N,\text{tr}}^n$ is the log principle Kirchoff tensor of $\mathbf{F^{n}_{N,\text{tr}}}$ and $\psi_V$ is the dissipation potential. We take a model where
\begin{equation}
    \psi_V\left(\tau_N\right)=\frac{1}{2 \nu_d}\left|\operatorname{dev}\left(\tau_N\right)\right|^2+\frac{1}{9 \nu_v}\left(\tau_N \cdot \mathbf{1}\right)^2
\end{equation}
where $\nu_d$ and $\nu_v$ are parameters controlling the dissipation of the deviatoric and dilational parts. In this work, we take them to be equal, representing the homogeneity and isotropicity of the material. This is merely for simplifying the simulation and can be released straightforwardly. We will not use them directly instead, we can put this equation back into Eq.\ref{append:viscupdate}, and get a version of this formula purely in terms of $\epsilon_N^{n+1}$ and $\epsilon_{N,\text{tr}}^n$
\begin{equation}
    \epsilon_N^{n+1} = A(\epsilon_{N,\text{tr}}^n-B\text{tr}(\epsilon_{N,\text{tr}}^n)\cdot \mathbf{1})
\end{equation}
In principle, we can write the parameters $A,B$ in terms of $\nu_d,\nu_v$, however since in our algorithm, all these parameters will be learnt from videos, we will directly learn parameter $A$ and $B$ in the update rule without bothering $\nu_d,\nu_v$. But it worth remembering our model comes from a dissipation potential.

\section{Ethical Statement}
We confirm that all data used in this paper for research and publication have been obtained and used in a manner compliant with ethical standards. The individuals engaged in all experiments have given consent for their use, or the data are sourced from publicly available datasets and were used in accordance with the terms of use and permissions. Furthermore, the publication and use of these data and models do not pose any societal or ethical harm. We have taken necessary precautions to ensure that the research presented in this paper respects individual rights, including the right to privacy and the fundamental principles of ethical research conduct.